\documentclass{article}

\usepackage{microtype}
\usepackage{graphicx}
\usepackage{subfigure}
\usepackage{booktabs} 
\usepackage{multirow}
\usepackage{hyperref}
\usepackage{url}

\usepackage[accepted]{icml2025}

\usepackage{amsmath}
\usepackage{amssymb}
\usepackage{mathtools}
\usepackage{amsthm}

\usepackage[capitalize,noabbrev]{cleveref}

\theoremstyle{plain}

\theoremstyle{definition}

\theoremstyle{remark}

\usepackage{xcolor}
\usepackage{enumitem}
\usepackage{float}
\usepackage{bm}
\usepackage{pifont}
\usepackage{gensymb}
\usepackage{xspace}
\newcommand{\eg}{\textit{e.g.}\xspace}
\newcommand{\ie}{\textit{i.e.}\xspace}

\newcommand{\ourmodel}{\textsc{CFP-Gen}}
\makeatletter
\newcommand{\thickhline}{%
    \noalign {\ifnum 0=`}\fi \hrule height 1pt
    \futurelet \reserved@a \@xhline
}
\makeatother

\usepackage[textsize=tiny]{todonotes}

\icmltitlerunning{CFP-Gen}

\begin{document}

\twocolumn[
\icmltitle{\ourmodel: Combinatorial Functional Protein Generation via \\ Diffusion Language Models}

\begin{icmlauthorlist}
\icmlauthor{Junbo Yin}{xxx,yyy,zzz}
\icmlauthor{Chao Zha}{xxx,yyy,zzz}
\icmlauthor{Wenjia He}{xxx,yyy,zzz}
\icmlauthor{Chencheng Xu}{xxx,yyy,zzz}
\icmlauthor{Xin Gao}{xxx,yyy,zzz}
\end{icmlauthorlist}

\icmlaffiliation{xxx}{Computer Science Program, Computer, Electrical and Mathematical Sciences and Engineering Division, King Abdullah University of Science and Technology (KAUST), Thuwal 23955-6900, Kingdom of Saudi Arabia}
\icmlaffiliation{yyy}{Center of Excellence for Smart Health (KCSH), KAUST}
\icmlaffiliation{zzz}{Center of Excellence for Generative AI, KAUST}

\icmlcorrespondingauthor{Xin Gao}{xin.gao@kaust.edu.sa}

\icmlkeywords{protein design, function}

\vskip 0.3in
]



\printAffiliationsAndNotice{Code and data available at https://github.com/yinjunbo/cfpgen.}  

\begin{abstract}

Existing PLMs generate protein sequences based on a single-condition constraint from a specific modality, struggling to simultaneously satisfy multiple constraints across different modalities. 
In this work, we introduce \textbf{\ourmodel}, a novel diffusion language model for \textbf{C}ombinatorial \textbf{F}unctional \textbf{P}rotein \textbf{GEN}eration. \ourmodel~facilitates the \textit{de novo} protein design by integrating multimodal conditions with functional, sequence, and structural constraints. Specifically, an Annotation-Guided Feature Modulation (AGFM) module is introduced to dynamically adjust the protein feature distribution based on composable {functional annotations}, \eg, GO terms, IPR domains and EC numbers. Meanwhile, the Residue-Controlled Functional Encoding (RCFE) module captures residue-wise interaction to ensure more precise control. Additionally, off-the-shelf 3D structure encoders can be seamlessly integrated to impose geometric constraints. We demonstrate that CFP-GEN enables high-throughput generation of novel proteins with functionality comparable to natural proteins, while achieving a high success rate in designing multifunctional proteins.

\end{abstract}

\section{Introduction}
\label{introduction}

\begin{figure}
\includegraphics[width=0.99\linewidth]{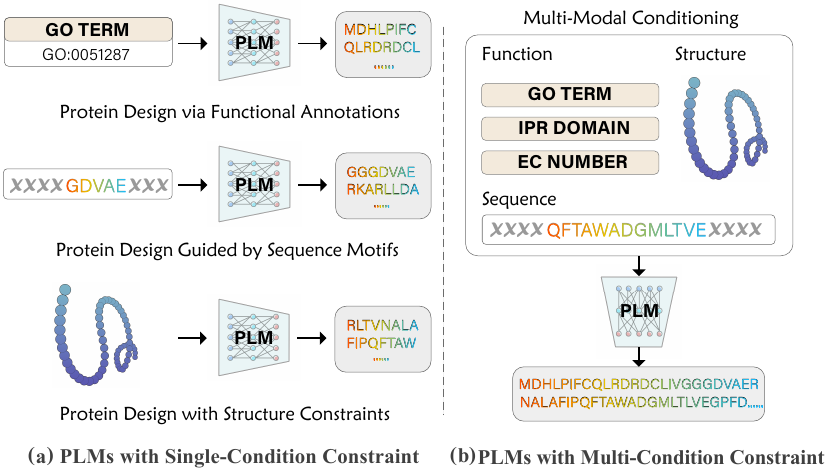}
\centering
 \caption{ \textbf{Motivation of \ourmodel.} (a) Previous PLMs typically generate proteins based on \textit{single-modality} conditioning, considering only individual functional constraints. (b) In contrast, \ourmodel~incorporates multiple conditions from \textit{diverse modalities}—function, sequence and structure—to impose comprehensive functional constraints, thereby leading to optimized proteins.} 
\label{fig:motivation}
\end{figure}

\textit{De novo} protein design~\cite{watson2023novo, krishna2024generalized,dauparas2022robust,wang2024diffusion, lisanza2024multistate} has emerged as a powerful strategy for numerous biotechnological applications, including drug development, enzyme engineering, and the creation of novel therapeutic proteins. The growing capabilities of large-scale protein language models (PLMs) offer tremendous potential for discovering novel proteins that do not exist in nature or would require millions of years of evolution to emerge~\cite{hayes2024simulating,madani2023large,alamdari2023protein}. 

Most current PLMs are primarily designed for unconditional protein generation~\cite{lin2023evolutionary,rencarbonnovo,qu2024p}, limiting their efficiency in addressing complex real-world challenges. 
In contrast, controllable protein generation~\cite{listov2024opportunities, ferruz2022controllable, madani2023large}, which tailors proteins to meet specific biological functions by explicitly defined conditions, holds significant promise for advancing practical applications.

Biologically meaningful proteins often simultaneously satisfy multiple functional constraints across diverse modalities, \eg, functional annotation, sequence or structure. As shown in Fig.~\ref{fig:motivation}, previous PLMs typically generate protein candidates based on a single-condition input from a specific modality, and thus have to rely on iterative filtering or multi-step optimization to meet constraints across multiple modalities~\cite{goverde2024computational}. Moreover, this pipeline often struggles to satisfy all desired functionalities and becomes impractical when available data is limited. Therefore, developing an advanced protein generative model capable of simultaneously handling multiple function constraints within a unified model would significantly streamline the multi-objective optimization in protein engineering.



In this work, we introduce 	\textbf{\ourmodel}, a novel large-scale diffusion language model specifically developed for \textbf{C}ombinatorial \textbf{F}unctional \textbf{P}rotein \textbf{GEN}eration. \ourmodel~iteratively denoises the input protein sequence while simultaneously accounting for various functional conditions, including functional annotations, desired sequence motifs and 3D atomic coordinates of reference proteins. 
During inference, composable annotation tags (\eg, GO terms, IPR domains and EC numbers) are encoded as one-hot embeddings and injected to the model through an Annotation-Guided Feature Modulation (AGFM) module. AGFM dynamically modulates the normalized feature distributions of the noised protein sequence by leveraging all available annotation tags. Compared to classifier-guided diffusion, this approach enables joint training to ensure strict alignment between functions and sequences, while also allowing for the flexible combinations of various functional annotations. 
Additionally, certain amino acid residues (\ie, sequence motif) are critical in real-world applications, as they often determine the desired functional properties. To cater to this need, a Residue-Controlled Functional Encoding (RCFE) module is introduced to explicitly encode these functional domains. RCFE employs an ESM-like transformer-based controller to effectively capture epistasis and evolutionary relationships among residues. As a result, this approach has the potential to generate novel protein sequences with optimized functional sites. Moreover, \ourmodel~supports direct optimization of user-provided proteins by encoding their 3D backbone atomic coordinates as conditional features using off-the-shelf structure encoders. By jointly incorporating functional constraints across diverse modalities, our approach ensures that the inverse folding process preserves structural accuracy while simultaneously satisfying functional requirements. This provides a significant advantage in tasks that demand multi-objective optimization.



Our contributions can be summarized as follows:
\vspace{-2mm}
\begin{itemize}
\item {A novel generative model, \ourmodel, is developed for \textit{de novo} design of functional proteins. \ourmodel~addresses multiple functional conditions across various modalities simultaneously, offering more promising starting points compared to previous PLMs limited by single-condition inputs.}
\vspace{-2mm}
\item  An Annotation-Guided Feature Modulation (AGFM) is introduced to enable highly-controllable protein generation by incorporating commonly used functional annotations in a composable and flexible manner.
\vspace{-2mm}
\item A Residue-Controlled Functional Encoder (RCFE) is proposed to encode functional domains at the residue level. It captures epistasis and interactions within sequences, enabling the generation of novel yet functional proteins.\end{itemize}

We thoroughly evaluate \ourmodel~across various tasks, including functional sequence generation, functional protein inverse folding, and multi-objective protein design. 
\ourmodel~achieves exceptional function performance, \eg, improving ESM3 by 30\% in $\textbf{\textit{F}}_{\text{1}}$-score, as demonstrated by leading function predictors. Additionally, it improves Amino Acid Recovery (AAR) of DPLM 
by 9\% in inverse folding.  
Notably, \ourmodel~demonstrates a remarkable success rate in designing multi-functional proteins (\eg, enzymes exhibiting multiple catalytic activities). We anticipate \ourmodel~to become a valuable and practical computational tool for addressing biomedical and biotechnological challenges.

\begin{figure*}
\includegraphics[width=0.99\linewidth]{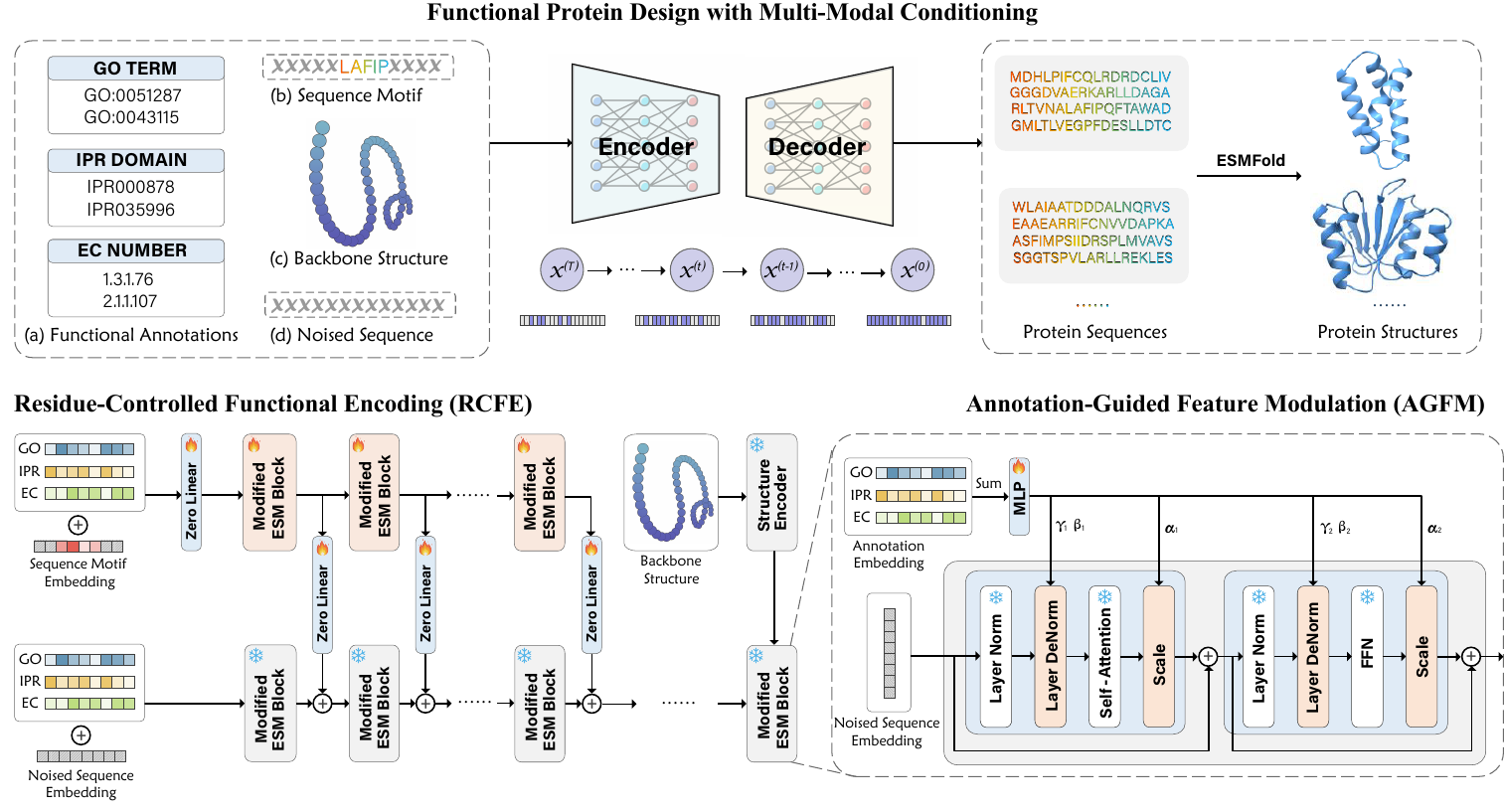}
\centering
 \caption{\textbf{Pipeline of~\ourmodel~model.} Functional conditions from diverse modalities, combined with the noised sequence, are iteratively processed by the model to generate desired proteins. Within each modified ESM block, AGFM adaptively adjusts the noised sequence embedding based on combinations of various functional annotations. Furthermore, sequence motifs and backbone atomic coordinates are embedded by RCFE and a structure encoder, respectively, providing precise and flexible guidance for the generation process.  
}
\label{fig:pipeline}
\end{figure*}

\section{Related Work}
\label{related_work}

PLMs have become indispensable tools in protein science, widely applied to both discriminative and generative tasks. ESM2~\cite{lin2023evolutionary} is a pioneering approach that utilizes scalable language models to uncover patterns in protein sequences across evolutionary space. By optimizing a masked modeling objective, ESM2 learns rich sequence representations, significantly improving performance in various discriminative tasks, including structure prediction~\cite{baek2021accurate} and functional classification~\cite{kulmanov2018deepgo}.
Meanwhile, ProtGPT2~\cite{ferruz2022protgpt2} performs unconditional sequence generation, akin to natural language sentence generation. 
Moving toward controllable sequence generation, ProGen~\cite{nijkamp2023progen2} augments the language model with protein family tags, enabling the generation of sequences within specific functional constraints.  
ProteoGAN~\cite{kucera2022conditional} incorporates GO terms for sequence generation, providing detailed \textit{in silico} validations of its output sequences.
Later, EvoDiff~\cite{alamdari2023protein} introduces a discrete diffusion model instead of the masked language modeling objective. This enables conditional protein generation, such as sequence inpainting and MSA-guided generation. 
Further, ZymCTRL~\cite{munsamy2022zymctrl} specializes in generating enzymes by conditioning on EC numbers, successfully designing sequences with catalytic activity. 
Most recently, DPLM~\cite{wang2024diffusion} is proposed as a new foundation model that not only supports representation learning for discriminative tasks but also enables conditional generation guided by secondary structure classifications and sequence or structure motifs.

All of the above PLMs are fundamentally limited to single-modality conditioning, \ie, generating sequences based on one functional constraint at a time. This reduces the flexibility and success rate of generated proteins when multiple functional constraints are required. Although the recently published ESM3~\cite{hayes2025simulating} supports multimodal inputs, it defines function using a limited vocabulary, restricting its broader applicability. In contrast, our model incorporates more diverse functions. This ensures more precise control and leads to functionally superior proteins.


\section{Methodology}
\label{sec:method}


We begin by introducing the essential concepts of \ourmodel~in \S\ref{subsec:framework}. Next, we present the AGFM module in \S\ref{subsec:agfm}. Following that, we detail the RCFE module in \S\ref{subsec:rcfe}, which is designed to encode desired sequence motifs. Finally, we demonstrate how \ourmodel~smoothly incorporates functional conditions across diverse modalities in \S\ref{subsec:inference}.

\subsection{Preliminaries of \ourmodel}
\label{subsec:framework}
Diffusion models are widely recognized for their capability in \textit{de novo} protein design. \ourmodel~is built upon the leading diffusion protein language model DPLM~\cite{wang2024diffusion} to leverage its well-pretrained parameters attained from evolutionary-scale datasets. Concretely, discrete diffusion mechanism~\cite{austin2021structured} is employed to model the protein sequence distribution at the amino acid category level.
Let $\mathbf{x} \sim q(\mathbf{x})$ denote a protein sequence of length $L$, represented as $\mathbf{x} = (x_1, x_2, \ldots, x_L)$, where each $x_i \in \{0, 1\}^{|V|}$ is a one-hot vector indicating an amino acid category from the set $V$ of 20 standard types. The categorical distribution $\text{Cat}(\mathbf{x}; \mathbf{p})$ models the sequence $\mathbf{x}$, with $\mathbf{p} = (\mathbf{p}_1, \mathbf{p}_2, \ldots, \mathbf{p}_L)$ as the collection of probability vectors. Each $\mathbf{p}_i = (p_{i,1}, p_{i,2}, \ldots, p_{i,|V|})$ specifies the categorical distribution of the $i$-th residue in the sequence, \ie, $p_{i,v}$ represents the probability of selecting amino acid category $v$, always ensuring that $\sum_{v \in V} p_{i,v} = 1, \forall i \in \{1, \ldots, L\}$.

\noindent\textbf{Forward Process with Discrete Diffusion.} 
We apply discrete diffusion to gradually corrupt the raw sequence $\mathbf{x}^{(0)}$ over $t \in \{1, \ldots, T\}$ time steps by transitioning each amino acid token towards a stationary noise distribution. 
The stationary noise distribution is parameterized by a fixed probability vector $\mathbf{q}_\text{noise}$ and can be expressed as:$
q_\text{noise}(\mathbf{x}^{(t)}) = \text{Cat}(\mathbf{x}^{(t)}; \mathbf{p} = \mathbf{q}_\text{noise})$.
Following DPLM, $q_\text{noise}(\mathbf{x}^{(t)})$ satisfies: $q_\text{noise}(\mathbf{x}^{(t)}) = 1$ if $\mathbf{x}^{(t)} = [X]$ and $0$ otherwise, where $[X]$ refers to the absorbing state (\eg, \texttt{<mask>}
). Once a token transitions into this absorbing state, it will remain unchanged in all subsequent diffusion steps. This ensures that the forward process eventually leads all amino acid tokens to \texttt{<mask>}, unifying the principles behind masked language models~\cite{kenton2019bert} and autoregressive language models~\cite{ferruz2022protgpt2}. 
Mathematically, the forward transition process is given by:
\begin{equation}
q(\mathbf{x}^{(t)} \mid \mathbf{x}^{(t-1)}) = \mathrm{Cat}(\mathbf{x}^{(t)}; \mathbf{p} = \mathbf{x}^{(t-1)}\mathbf{Q}_t),	
\end{equation}
where $\mathbf{Q}_t$ is the transition matrix at step $t$. Each row of $\mathbf{Q}_t$ is a probability vector defined as:
$
\mathbf{Q}_t = \beta_t I + (1 - \beta_t) \mathbf{q}_{\text{noise}},
$
where $I$ is the identity matrix and $\beta_t \in [0, 1]$ is the noise schedule.
Due to the Markov property, the overall transition from \(\mathbf{x}^{(0)}\) to \(\mathbf{x}^{(t)}\) can be expressed as follows:
\begin{equation}
q(\mathbf{x}^{(t)} \mid \mathbf{x}^{(0)}) = \text{Cat}(\mathbf{x}^{(t)}; \mathbf{p} = \alpha_t \mathbf{x}^{(0)} + (1 - \alpha_t) \mathbf{q}_{\text{noise}}),
\end{equation}
where \(\alpha_t = \prod_{i=1}^t \beta_i\) represents the accumulated effect of the noise schedule over \(t\) steps. As \(t \to T\), \(\alpha_t \to 0\), ensuring that the sequence \(\mathbf{x}^{(t)}\) converges to the stationary noise \(\mathbf{q}_{\text{noise}}\) at time step $T$.

\noindent\textbf{Reverse Denoising with Composable Conditions.} 
To enable the generation of protein sequences with desired functionalities, we incorporate multimodal conditions into the discrete diffusion framework during the reverse process, as illustrated in Fig.~\ref{fig:pipeline}. These include 0D annotation tags $c_{\text{anno}}$, 1D sequence motifs $c_{\text{seq}}$, and 3D structures $c_{\text{str}}$, encoded by \(f_{\text{AGFM}}\), \(f_{\text{RCFE}}\), and \(f_{\text{GVPT}}\), respectively. The details of these networks will be introduced in the following sections, where GVPT indicates the GVP-Transformer~\cite{hsu2022learning} for encoding 3D backbone atomic coordinates.

The reverse process reconstructs the sequence by iteratively denoising \(\mathbf{x}^{(t)}\) back to \(\mathbf{x}^{(0)}\), using the predicted \(\hat{\mathbf{x}}^{(0)}\) at each step, which is derived from the KL divergence, \ie, $D_\mathrm{KL}\left[q(\mathbf{x}^{(t-1)} \mid \mathbf{x}^{(t)}, \mathbf{x}^{(0)}) \parallel p_\theta(\mathbf{x}^{(t-1)} \mid \mathbf{x}^{(t)})\right]$. Thus, the reverse step is expressed as follows:
\begin{equation}
p_\theta(\mathbf{x}^{(t-1)} \mid \mathbf{x}^{(t)}) = \sum_{\hat{\mathbf{x}}^{(0)}} q(\mathbf{x}^{(t-1)} \mid \mathbf{x}^{(t)}, \hat{\mathbf{x}}^{(0)}) p_\theta(\hat{\mathbf{x}}^{(0)} \mid \mathbf{x}^{(t)}, c),
\end{equation}
where \( c \in \{c_{\text{anno}}, c_{\text{seq}}, c_{\text{str}}\} \) represents any combinations of conditions, depending on their availability. To predict \(\hat{\mathbf{x}}^{(0)}\), the model incorporates \(c\) throughout the network:
\[
p_\theta(\hat{\mathbf{x}}^{(0)} \mid \mathbf{x}^{(t)}, c) = \text{Softmax}(W h(\mathbf{x}^{(t)}, c)),
\]
\begin{equation}
\label{eq:main}
h(\mathbf{x}^{(t)}, c) = \mathcal{A} \Big( 
    f_\text{RCFE} \big(f_\text{AGFM}(\mathbf{x}^{(t)}, c_{\text{anno}}), c_{\text{seq}} \big), 
    f_{\text{GVPT}}(c_{\text{str}})
\Big),
\end{equation}
where  \(W\) is the final output layer, \(\mathcal{A}\) is a cross-attention layer and $h(\mathbf{x}^{(t)}, c)$ represents the core of \ourmodel, sequentially integrating modules such as \(f_{\text{AGFM}}\), \(f_{\text{RCFE}}\) and \(f_{\text{GVPT}}\). 

\textbf{Optimization of \ourmodel.} The training objective is to optimize the predicted \(\hat{\mathbf{x}}^{(0)}\) against the raw sequence \(\mathbf{x}^{(0)}\) using a weighted cross-entropy loss:
\begin{equation}
\label{eq:gamma}
\mathcal{L} = \mathbb{E}_{q(\mathbf{x}^{(0)})} \left[\lambda^{(t)} \sum_{1 \leq i \leq L} b_i(t) \cdot \log p_\theta(\mathbf{x}_i^{(0)} | \mathbf{x}^{(t)}, \gamma(c)) \right].
\end{equation}
Here, \(\lambda^{(t)}\) adjusts the influence of each diffusion time step \(t\), \(b_i(t)\) determines the contribution of each position \(i\), and \(\gamma(c)\) controls the strength of the condition \(c\).
In the following sections, we demonstrate how the inclusion of $c$ facilitates the prediction of highly functional proteins and how its composable nature empowers \ourmodel~to achieve multi-objective protein design.

\subsection{Annotation-Guided Conditioning by AGFM}
\label{subsec:agfm}
Functional annotations are both highly informative and representative, as they are rigorously curated by biological experts to capture essential properties of proteins. In this work, we consider three commonly used annotations, \ie, GO terms~\cite{gene2021gene}, IPR numbers~\cite{hunter2009interpro}, and EC numbers~\cite{schomburg2004brenda}, to guide the generation process. Typically, each type of annotation characterizes the protein's profile from a different perspective: GO captures the molecular functions, IPR defines the functional domains, and EC describes the related catalytic processes. Unlike ESM3~\cite{hayes2025simulating} relying only on a restricted vocabulary mapping to represent IPR annotations, we adopt a more tactful approach to combine diverse functions. While three types of annotations are presented as examples, our paradigm is extensible to other forms of annotations, such as Pfam~\cite{bateman2000pfam}.

Existing PLMs often encode a single type of annotation. For instance, ZymCTRL~\cite{munsamy2022zymctrl} designs enzymes conditioned solely on EC numbers, while ProteoGAN~\cite{kucera2022conditional} operates only with GO terms. Although ProGen~\cite{madani2023large} supports multiple annotation tags, it requires extensive fine-tuning with sufficient homologous sequences to ensure performance, which limits its application when training samples are scarce. By contrast, \ourmodel~is more flexible and versatile by addressing multiple annotations simultaneously, without the need of further fine-tuning. By leveraging the complementary characteristics of these high-quality annotations, we also enable a more comprehensive description of a protein.

As shown in Fig.~\ref{fig:pipeline}, an Annotation-Guided Feature Modulation (AGFM) module is integrated into each modified ESM block~\cite{lin2023evolutionary}, where each block has been well pre-trained by DPLM. \ourmodel~is achieved by training only the annotation embedding layers and a single MLP layer.
 For each type of annotation (\eg, GO, IPR, EC), we maintain a dedicated embedding layer. Each annotation is mapped to a same-dimensional vector representation through its respective embedding layer. This flexible design ensures that annotations from different sources can be directly summed together. The process is formally represented as:
$\mathbf{x}_\text{anno} = f_\text{AGFM}(\mathbf{x}, \mathbf{c}_\text{GO}, \mathbf{c}_\text{IPR}, \mathbf{c}_\text{EC})$, where $\mathbf{x}, \mathbf{x}_\text{anno} \in \mathbb{R}^{L \times D}$ denotes the input feature and the modulated feature output by AGFM within one block, based on the denoised sequence $\mathbf{x}^{(t)}$ at time step $t$. Concretely, the sum of embeddings is passed through an MLP layer to regress three parameters, \ie, 
$\boldsymbol{\gamma}, \boldsymbol{\beta}, \boldsymbol{\alpha} =\mathcal{F}(\mathbf{c}_\text{GO} + \mathbf{c}_\text{IPR} + \mathbf{c}_\text{EC})$,
where \(\boldsymbol{\gamma}\), \(\boldsymbol{\beta}\) and $\boldsymbol{\alpha}$ are the scaling, shifting and gating factors, respectively.
The MLP layer \(\mathcal{F}\) follows a specialized initialization strategy: the weights predicting \(\boldsymbol{\gamma}\) and \(\boldsymbol{\beta}\) are initialized to zeros, ensuring that conditional information is gradually injected into \(\mathbf{x}^{(t)}\), while the weights predicting \(\boldsymbol{\alpha}\) are initialized to ones, allowing for effective gating from the beginning.
The scale \(\boldsymbol{\gamma}\) and shift \(\boldsymbol{\beta}\) parameters modulate the distribution of the noised feature \(\mathbf{x}\) before both the self-attention (SA) and feed-forward network (FFN) layers within each ESM block, akin to a feature de-normalization process: $\mathbf{x}_\text{out} = \boldsymbol{\gamma} \odot \mathbf{x} + \boldsymbol{\beta}$.
Additionally, the gating factor \(\boldsymbol{\alpha}\) operates on the feature distribution \(\mathbf{x}_\text{out}^\prime\) output by the SA and FFN layers through the following formulation:
$\hat{\mathbf{x}}_\text{out} = \boldsymbol{\alpha} \odot \mathbf{x}_\text{out}^\prime + \mathbf{x}_\text{out}^\prime$.



Through the collaboration of diverse annotations, AGFM effectively adjusts the intermediate representations during diffusion, resulting in higher-quality proteins compared to previous approaches relying on single-condition control.

\subsection{Sequence-Guided Conditioning with RCFE}
\label{subsec:rcfe}
While AGFM has provided effective control over protein generation with annotations, certain applications require finer-grained guidance, \eg, biological experts often focus on functional sequence fragments (referred to as sequence motifs) to ensure that the generated proteins retain desired functionality associated with these motifs. Thus, we propose the Residue-Controlled Functional Encoder (RCFE) with an additional network branch to handle this sequence-level condition. The core of RCFE lies in inferring the complete sequence while preserving or even optimizing the specified sequence motifs, akin to the sequence inpainting task introduced by DPLM~\cite{wang2024diffusion} and EvoDiff~\cite{alamdari2023protein}. However, unlike these methods, which rely on fixed amino acids throughout generation, RCFE dynamically updates the specified motifs during inference. This dynamic adjustment enables the model to emulate evolutionary processes, offering the potential to discover improved sequence motifs with enhanced functional properties.

Specifically, $f_\text{RCFE}$ in Eq.\ref{eq:main} consists of two branches of modified ESM blocks, along with zero-initialized linear layers to jointly handle sequence-level and annotation-level information. As shown in Fig.\ref{fig:pipeline}, the bottom gray blocks represent the main branch, which processes the noised sequence and annotation tags, while the upper orange blocks represent the second branch, composed of trainable copies of the main branch, for tackling sequence-level conditioning. Inspired by ControlNet~\cite{zhang2023adding}, RCFE adapts this dual-branch design to transformer-based ESM blocks, making it well-suited for protein sequence generation. 

Let $\mathcal{E}_\text{esm}(\cdot, \Theta_\text{esm})$ denote a modified ESM block in the main branch, which has been enhanced by AGFM to integrate the functional annotations. The parameters $\Theta_\text{esm}$ in the main branch blocks remain frozen during training RCFE, while $\Theta_{\text{seq}}$ in another branch blocks using a trainable copy of $\Theta_\text{esm}$ need to be updated, denoted as $\mathcal{E}_\text{seq}(\cdot, \Theta_{\text{seq}})$. This design preserves the powerful representation capabilities of the main branch that has been fully trained, while enabling the other branch to dynamically encode sequence motifs. Formally,
we first pad the sequence motif with \texttt{<mask>} and project it into the same latent space as the denoised feature \( \mathbf{x} \in \mathbb{R}^{L \times D} \) within one block, which is denoted as \( \mathbf{c}_\text{seq} \in \mathbb{R}^{L \times D}\). In this way, the workflow of RCFE to handle the sequence-level condition can be represented as follows:
\begin{equation}
\label{eq:control}
\mathbf{x}_\text{seq} = \mathcal{E}_\text{esm}(\mathbf{x}; \Theta_\text{esm}) + \mathcal{F}_\text{out}(\mathcal{E}_{\text{seq}}(\mathbf{x} + \mathcal{F}_\text{in}(\mathbf{c}_\text{seq}; \Theta_\text{in}); \Theta_{\text{seq}}); \Theta_\text{out}),
\end{equation}
where $\mathcal{F}_\text{in}(\cdot; \Theta_\text{in})$ denotes a zero-initialized linear layer applied to $\mathbf{c}_\text{seq}$. Note that this layer is only utilized in the first block of the conditional branch. Similarly, $\mathcal{F}_\text{out}(\cdot; \Theta_\text{out})$ is another zero-initialized linear layer. Zero-initialization enables the gradual incorporation of meaningful information from the sequence motif condition while ensuring the stability of the main blocks. Finally, $\mathbf{x}_\text{seq}$ represents the updated feature from a main block, enriched with conditional information at both the annotation and sequence motif levels. It is worth mentioning that, in image generation models, typically only the encoder part of a U-Net is used as the trainable branch. Similarly, we leverage only the first half blocks of the ESM2 model to encode sequence motifs, ensuring an efficient yet expressive representation.

Consequently, the sequence-level condition serves as a strong complement to the annotation-level condition, leading to enhanced controllability over the generated sequences compared to using single condition modality alone.

\subsection{Multimodal Conditioning within \ourmodel}
\label{subsec:inference}

\begin{table*}[ht]
    \centering
    \caption{\textbf{Function evaluation of the generated sequences using diverse function predictors}, \eg, DeepGO-SE~\cite{kulmanov2024protein} for GO terms, InterProScan~\cite{blum2021interpro} for IPR domains, and CLEAN~\cite{yu2023enzyme} for EC numbers. The best result in each column is highlighted in \textbf{bold}, while the second-best is \underline{underlined} (excluding the positive control).  }
       \resizebox{0.98\textwidth}{!}
    {
    \begin{tabular}{l|c|c|c|c|c|c|c|c}
        \toprule
        \textbf{Models} & \textbf{Supported Condtions} & \textbf{MRR}$\uparrow$ & \textbf{MMD}$\downarrow$  & \textbf{MMD-G}$\downarrow$   & \textbf{mic. $F_{\text{1}}$}$\uparrow$ & \textbf{mac. $F_{\text{1}}$}$\uparrow$ & \textbf{AUPR}$\uparrow$ &  \textbf{AUC}$\uparrow$ \\
        \midrule
        \multicolumn{9}{c}{\textbf{\textit{Evaluating the Protein Functionality via predicted GO Terms.}}} \\
        \midrule
       Positive Control & - & 0.939 & 0.000 &0.000 & 0.543 & 0.522 & 0.402 & 0.775  \\
       Negative Control & - & 0.017 & 0.215 & 0.125 & 0.205 & 0.020 & 0.025 & 0.501 \\
          DPLM~\cite{wang2024diffusion} & Sequence Motif, Structure& 0.134 & 0.189& 0.109 & 0.332& 0.189 & 0.109 & 0.581  \\
     ProteoGAN~\cite{kucera2022conditional} & GO Term & 0.277 & 0.095 & 0.055 & 0.376 & 0.093 & 0.121 & 0.510  \\

       ProGen2~\cite{nijkamp2023progen2} & GO Term, Sequence Motif & 0.545 & 0.109 & 0.064 & 0.414 & 0.355  & 0.240 & 0.663 \\
           
        \midrule
        \ourmodel~\textit{(w/ GO)}   & GO, IPR, EC, Seq., Struc. & 0.601 & 0.112 & 0.060 & 0.429& 0.370 & 0.245& 0.674   \\
       \ourmodel~\textit{(w/ GO and IPR)}  & GO, IPR, EC, Seq., Struc.  & 0.779 & 0.066 & 0.039 &0.496 & {0.458} & 0.339& 0.732 \\
        \ourmodel~\textit{(w/ Motif)}  & GO, IPR, EC, Seq., Struc.  & \underline{0.839} & \underline{0.046} &\underline{0.028} & \underline{0.504} & \underline{0.492} & \underline{0.370} & \underline{0.762} \\
       \ourmodel~\textit{(w/ GO, IPR and Motif)} &  GO, IPR, EC, Seq., Struc. & \textbf{0.870} & \textbf{0.036}& \textbf{0.022} & \textbf{0.532}& \textbf{0.550} & \textbf{0.435}  &\textbf{0.795}\\

           \midrule
         \multicolumn{9}{c}{\textbf{\textit{Evaluating the Protein Functionality via predicted IPR Domains.}}} \\
        \midrule
        
 Positive Control & - & 1.000	 & 0.000 & 0.000 & 1.000 &1.000 & 1.000 & 1.000  \\
       Negative Control & - & 0.016  & 0.224 & 0.131 & 0.003 & 0.000 & 0.004 &0.500 \\
        DPLM~\cite{wang2024diffusion} &  Sequence Motif, Structure & 0.053 & 0.192 & 0.112 & 0.548 & 0.422 & 0.369 & 0.695   \\
     ESM3~\cite{hayes2025simulating}& IPR Domain, Seq., Struc.  & 0.148 & 0.101 & 0.060 & 0.690 & 0.565  & 0.502 & 0.762  \\
      ProGen2~\cite{nijkamp2023progen2} & GO Term, Sequence Motif& 0.281 & 0.112 & 0.067 & 0.712& 0.584 & 0.536 & 0.772  \\
      
        \midrule
        \ourmodel~\textit{(w/ IPR)}   &  GO, IPR, EC, Seq., Struc.  & 0.332 & 0.094 & 0.056 & 0.882& 0.826 & 0.782 & 0.899   \\
               \ourmodel~\textit{(w/ GO and IPR)} &  GO, IPR, EC, Seq., Struc.  & 0.386 & 0.078 & 0.047 & 0.909& 0.858 & 0.824  & 0.922\\
          \ourmodel~\textit{(w/ Motif)}  & GO, IPR, EC, Seq., Struc.  & \underline{0.583} & \underline{0.056} & \underline{0.035} & \underline{0.937} & \underline{0.927}  & \underline{0.891} & \underline{0.970} \\ 
       \ourmodel~\textit{(w/ GO, IPR and Motif)} &  GO, IPR, EC, Seq., Struc.  & \textbf{0.654} & \textbf{0.046} & \textbf{0.030} & \textbf{0.983} & \textbf{0.975} & \textbf{0.965}  &\textbf{0.992} \\
        \midrule
         \multicolumn{9}{c}{\textbf{\textit{Evaluating the Protein Functionality via predicted EC Numbers.}}} \\
        \midrule
     Positive Control & -& 1.000 & 0.000 & 0.000 & 0.943 & 0.927 & 0.895 & 0.952   \\
       Negative Control & - & 0.012 & 0.222 & 0.129  & 0.036 & 0.028 & 0.016 & 0.510  \\
          DPLM~\cite{wang2024diffusion} & Sequence Motif, Structure & 0.211 & 0.189 & 0.109 & 0.241 & 0.186 & 0.127 & 0.572  \\
   ProGen2~\cite{nijkamp2023progen2} & GO Term, Sequence Motif& 0.530 & 0.127 & 0.075 & 0.519&0.387 & 0.318 & 0.661   \\
       ZymCTRL~\cite{munsamy2022zymctrl} & EC Number & 0.562 &\underline{0.044} & \underline{0.026} & 0.901 & 0.774  &0.743& 0.876 \\
      
        \midrule
        \ourmodel~\textit{(w/ EC)}   &  GO, IPR, EC, Seq., Struc.  & 0.567  & 0.141 & 0.082 & 0.722 &0.673& 0.559 & 0.786  \\
            \ourmodel~\textit{(w/ EC, GO and IPR)} &  GO, IPR, EC, Seq., Struc.  & 0.774 & 0.103 & 0.060 & 0.780 &0.743 &0.644 &0.831\\
       \ourmodel~\textit{(w/ EC and Motif)} &  GO, IPR, EC, Seq., Struc.  &\underline{0.898} & 0.049 & 0.029 & \underline{0.937}& \underline{0.920} &\underline{0.883} & \underline{0.946} \\
       \ourmodel~\textit{(w/ EC, GO, IPR and Motif)}   &  GO, IPR, EC, Seq., Struc.  & \textbf{0.924} & \textbf{0.041} & \textbf{0.024} &\textbf{0.942} & \textbf{0.925}& \textbf{0.892}& \textbf{0.951}   \\

        \bottomrule
    \end{tabular}
    }
         \label{tab:sota_comparison}
\end{table*}

The fundamental principle in protein science is that function, sequence, and structure are inherently interdependent. This intricate relationship reveals the importance of jointly modeling these modalities to achieve accurate protein design. 

To this end, we consider a practical scenario where biological experts have obtained the backbone structures of interest (\eg, from RFDiffusion~\cite{watson2023novo}) and aim to determine the functional sequence that folds into this structure, a problem known as the inverse folding task. In the traditional paradigm, the goal is to generate a sequence that correctly folds into a given backbone structure. However, when additional functional constraints \( {c}_\text{anno} \) are required, this task can be extended to \textit{functional protein inverse folding} or the \textit{inverse function task} - to generate a sequence \( \mathbf{x}^* \) that maximizes sequence recovery while simultaneously optimizing the functionality recovery matching \( {c}_\text{anno} \):
\begin{equation}
\label{eq:func}
\mathbf{x}^* = \arg\max_{\mathbf{x}} p(\mathbf{x} \mid {c}_\text{anno}, {c}_\text{seq}, {c}_\text{str}) + S_{\text{func}}(\mathbf{x}, {c}_\text{anno}),
\end{equation}
where $S_\text{func}$ is a scoring function that evaluates how well the generated sequence \( \mathbf{x} \) aligns with the desired functional properties. More discussion on  \( S_{\text{func}} \) can be found in \S\ref{subsec:evaluation}.

To achieve the objective in Eq.~\ref{eq:func}, we employ a structure encoder \( f_{\text{GVPT}} \) (\ie, GVP-Transformer) to embed the backbone atom coordinates of the protein: $\mathbf{c}_\text{str} = f_{\text{GVPT}}(c_\text{str})$.
Subsequently, \( \mathbf{c}_\text{str} \) is injected into the final ESM block of the main branch through a cross-attention layer, following the strategy of DPLM. In our implementation, we found that the pretrained cross-attention layer from DPLM could be directly utilized without the need of further training, suggesting that more functional adapters can be incorporated.  
The denoised features, having already incorporated both \( \mathbf{c}_\text{anno} \) and \( \mathbf{c}_\text{seq} \), further interact with \( \mathbf{c}_\text{str} \) to enable a more precise and functionally coherent protein design.  

Our functional protein inverse folding reduces the search space from a vast sequence space to sequences that are both structurally viable and functionally relevant. Compared to traditional inverse folding, which rely only on structural constraints, our approach incorporates functional information as an additional constraint. This leads to higher sequence recovery rates. We demonstrate significant improvements for both the zero-shot and supervised fine-tuning (SFT) versions of \ourmodel~with structural adapter in \S\ref{sec:inverse}.

\section{Experiments} 
\label{experiments}

\subsection{Experimental Setup} 
\label{subsec:setup}
\noindent\textbf{Datasets.} To collect high-quality data for training \ourmodel, we employ expert-curated functional annotations from SwissProt (UniProtKB)~\cite{uniprot2019uniprot}, InterPro~\cite{hunter2009interpro}, and CARE~\cite{yang2024care} databases. Two datasets were constructed for general protein design and enzyme design, respectively. For the \textit{general protein dataset}, we include 103,939 protein sequences covering 375 GO terms and 1,154 IPR domains. For the \textit{enzyme design dataset}, we intersected SwissProt with the CARE dataset, yielding 139,551 enzyme sequences annotated with 661 EC numbers (4-level EC annotations).  
Additionally, the PDB~\cite{berman2000protein} and AFDB~\cite{jumper2021highly} databases were exploited to provide backbone atomic coordinates, ensuring structural constraints are incorporated into the dataset. 
Additional details regarding the datasets can be found in the Appendix \S\ref{subsec:dataset}.

\noindent\textbf{Implementation Details.} Our model is built upon the pretrained DPLM-650M and trained in two progressive stages. 
First, we train the AGFM module to enable effective integration of functional annotations. 
The pretrained parameters from the first stage are then used in the next stage, where we train the copy ESM blocks in RCFE, allowing the model to condition on sequence motifs. The optimization schedule is the same as unconditional DPLM.
For the structural conditioning, we directly use the GVP-Transformer and structure adapter from DPLM, which are pretrained on the CATH~\cite{sillitoe2021cath} database, without additional fine-tuning for \ourmodel.  
During inference, users can specify any functional constraints and their combinations as conditioning signals. Further model implementation details can be found in the Appendix \S\ref{subsec:details}.

\begin{table}[t]
    \centering
    
     \caption{\textbf{Evaluation of functional protein inverse folding.} \ourmodel~achieves significantly higher AAR and improved functionality compared to other inverse folding methods.}
       \resizebox{0.5\textwidth}{!}
    {
    \begin{tabular}{l|c|c|c|c|c}
        \toprule
        \textbf{Models} & \textbf{AAR} & \textbf{MRR}  & \textbf{$F_{\text{max}}$} & \textbf{scTM} & \textbf{pLDDT} \\
        \midrule
 
    Positive Control & 100.00 & 0.939 &0.574 & 0.910 & 86.12 \\
      ProteinMPNN~\cite{dauparas2022robust} & 45.28 & 0.133 & 0.470 & \textbf{0.902} & 85.04 \\
        ESM-IF~\cite{hsu2022learning} & 57.39 & 0.495 & 0.551 & 0.895 & 83.49 \\
        LM-DESIGN~\cite{zheng2023structure} &63.81 & 0.730 & 0.549 & \underline{0.898} & \textbf{85.46} \\
        DPLM~\cite{wang2024diffusion} & 66.94 & 0.721 & 0.552 & 0.883 & \underline{85.33} \\
          \midrule
       \ourmodel~{(\text{Zero-Shot}, \textit{w/ GO and IPR})}  & \underline{72.05} & \underline{0.866} & \underline{0.571} & 0.887 & 83.28 \\
        \ourmodel~{(\text{SFT}, \textit{w/ GO and IPR})}  & \textbf{76.39} & \textbf{0.882} & \textbf{0.581} & 0.889 & 83.53 \\

        \bottomrule
    \end{tabular}
    }
 
    \label{tab:inverse}
\end{table}

\subsection{Benchmarking Protein Functional Performance}
\label{subsec:evaluation}
In this section, we evaluate the functional properties of the generated protein sequences. To achieve this, we split two validation datasets from the general protein dataset and enzyme dataset introduced in \S\ref{subsec:dataset}. We evaluate GO and IPR functions on the \textit{general protein dataset} and EC function on the \textit{enzyme design dataset}. The functional annotations of each natural sequence are used as conditioning prompts for the PLMs, generating one sequence per prompt with varying length to strictly evaluate nearest-neighbor performance. 
We assess the similarity between the generated and the real sequences using various metrics, such as Maximum Mean Discrepancy (MMD), including its Gaussian kernel variant (MMD-G), and Mean Reciprocal Rank (MRR). Then, various function predictors (\ie, $S_{\text{func}}$ in Eq.~\ref{eq:func}) are applied to assign functional labels and evaluate whether the generated sequences exhibit functional consistency. Specifically, we use DeepGO-SE~\cite{kulmanov2024protein} for GO label prediction, InterProScan~\cite{blum2021interpro} for homology-based annotation, and CLEAN~\cite{yu2023enzyme} for catalytic function prediction. Since these predicted labels correspond to a multi-label classification problem, we adopt commonly used function prediction metrics~\cite{kim2023functional} to quantitatively evaluate the alignment between predicted functions and the prompt functions (\ie, ground-truths), such as micro $\textbf{\textit{F}}_{\text{1}}$-score, macro $\textbf{\textit{F}}_{\text{1}}$-score, macro AUPR and macro AUC. Details about the evaluation metrics are in Appendix \S\ref{subsec:metrics}.

We conduct performance comparison against leading PLMs under varying condition guidance. Notably, most of these PLMs support only single-condition input, whereas \ourmodel~allows combinations of input conditions across different modalities. Detailed implementation for sequence generation using each approach can be found in the Appendix \S\ref{subsec:plms}. Furthermore, we provide positive and negative controls to establish the upper and lower bounds of these evaluation metrics. The positive control consists of the real sequences from UniProtKB, while the negative control is the unconditionally generated sequences by DPLM. 

\begin{figure}
\includegraphics[width=0.95\linewidth]{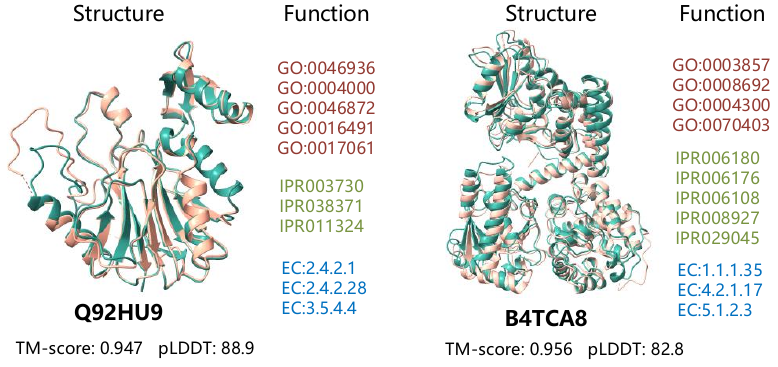}
\centering
 \caption{ \textbf{Examples of multi-catalytic enzymes.} \ourmodel~generates high-quality proteins (\ie, TM-score above 90) with multimodal conditions. The ground-truth structures from the AFDB database are in green, while the generated structures are in red.}
\label{fig:proteins}
\end{figure}

\subsubsection{Evaluation of GO terms}
As shown in Tab.~\ref{tab:sota_comparison}, the sequences in the negative control exhibit little functional consistency with the prompt GO terms (macro $\textbf{\textit{F}}_{\text{1}}$-score = 0.02), proving the necessity of conditional generation. While both DPLM and ProteoGAN generate some functional proteins, ProGen2 achieves improved function when conditioned on the first 30 residues. In contrast, \ourmodel~demonstrates superior performance across various conditioning levels.
Firstly, the generated sequences closely match the distribution of real sequences, as evidenced by low MMD. Even when conditioned only on GO terms, the model attains satisfactory functional consistency. Further integrating IPR information enhances performance, yielding a micro $\textbf{\textit{F}}_{\text{1}}$-score of 0.496 and a macro $\textbf{\textit{F}}_{\text{1}}$-score of 0.458.
To further assess the impact of sequence motifs, we extract sequence fragments from IPR annotations, selecting those whose descriptions best align with the corresponding GO functions (resulting in 10–30 residues). Notably, even when using only these residues as conditioning inputs, the model generates functional sequences, reinforcing the strong correlation between sequence and function.
By combining sequence motifs and functional annotations, the designed sequences outperform real sequences (positive control), achieving a macro $\textbf{\textit{F}}_{\text{1}}$-score of 0.550 $\textit{vs.}$ 0.522, macro AUPR of 0.435 $\textit{vs.}$ 0.402, and AUC of 0.795 $\textit{vs.}$ 0.775. These results highlight the effectiveness of \ourmodel~in designing functionally enriched proteins.

\subsubsection{Evaluation of IPR Domains}
The evaluation results based on IPR domains provide further evidence of the superior performance of \ourmodel. Since running InterProScan on large-scale sequences is computationally expensive, we perform the evaluation on a uniformly downsampled subset, using one-tenth of the GO validation set.  
We also implement the recently published ESM3, which supports IPR domain site input. Since ESM3 is a masked prediction model, we provide an additional 30 residue from the real sequences as prompts. Similar residue prompts are also given to DPLM and ProGen2 for sequence inpainting. ProGen2 demonstrates improved functionality. However, \ourmodel~demonstrates strong performance in terms of MRR and micro/macro $\textbf{\textit{F}}_{\text{1}}$-score without requiring any sequence-level prompts. When further incorporating sequence motifs, it achieves functionality comparable to that of natural proteins, \ie, an macro $\textbf{\textit{F}}_{\text{1}}$-score of 0.975.

\begin{figure} 
\includegraphics[width=0.99\linewidth]{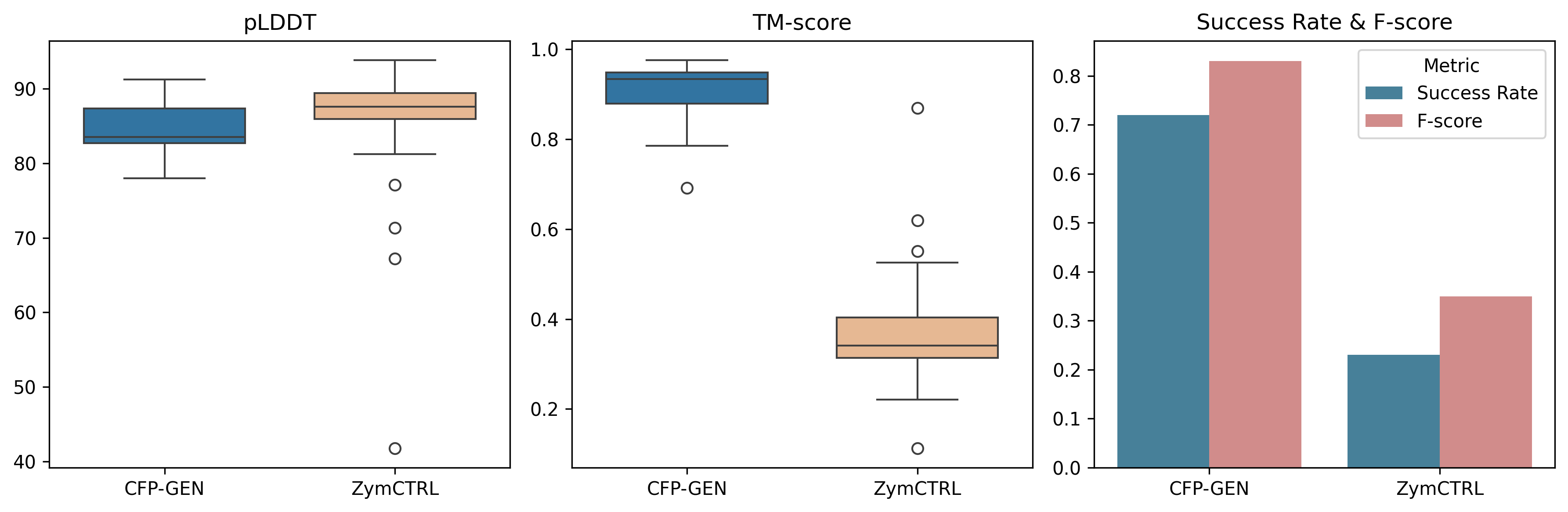}
\centering
 \caption{ \textbf{Evaluation of multi-catalytic enzyme design.} Our generated proteins exhibit high designability, meanwhile achieving high success rate and functionality as validated by CLEAN.}
\label{fig:multi-ec}
\end{figure}

\subsubsection{Evaluation of EC Numbers}
In this section, we mainly compare \ourmodel~with ZymCTRL. Please note that some EC numbers are not included in ZymCTRL and we thus evaluate it on a subset, where ZymCTRL achieves slightly higher results. Additionally, we provide sequence segments derived from IPR annotations as input prompts to DPLM and ProGen2 to conduct the sequence inpainting task for further comparison. 
As expected, the natural sequences in the positive control exhibit strong enzymatic activity, validating CLEAN as a reliable enzyme function predictor. \ourmodel, when conditioned only on EC numbers, generates sequences with high catalytic activity. When further incorporating GO and IPR annotations, we observe significant improvements in both distribution-level metrics (\ie, improving 21\% in MRR) and macro $\textbf{\textit{F}}_{\text{1}}$-score (\ie, improving 7\%), highlighting the importance of integrating comprehensive functional annotations. Finally, combining both functional labels and sequence motifs achieves the highest functional scores, approaching that of natural ones (\ie, 0.925 \textit{vs.} 0.924 in macro $\textbf{\textit{F}}_{\text{1}}$-score). These results further prove the effectiveness of \ourmodel.

\subsection{Functional Protein Inverse Folding}
\label{sec:inverse}
In Tab.~\ref{tab:inverse}, we evaluate the inverse folding task on the general protein dataset, where backbone structure information, combined with the GO and IPR annotations are provided as input. Several state-of-the-art methods, including ProteinMPNN, ESM-IF, LM-DESIGN and DPLM trained on CATH are used for comparison.  
Notably, most of these methods—including \ourmodel~(Zero-Shot)—employ encoder networks pretrained only on UniProtKB. In particular, \ourmodel~(Zero-Shot) does not involve any fine-tuning of DPLM’s structural adapter. For comparison, we also provide a supervised fine-tuned variant, \ourmodel~(SFT), in which the structural adapter is trained on our general protein dataset using backbone structure inputs.
The generated sequences are then folded using ESMFold~\cite{lin2023evolutionary} to assess their designability.  
Overall, most methods achieve high $\textbf{\textit{F}}_{\text{max}}$-score (used in DeepGO-SE~\cite{kulmanov2024protein}), revealing the strong coupling between structure and function. Moreover, LM-DESIGN and DPLM demonstrate superior AAR and MRR performance. \ourmodel~(SFT), guided by functional annotations, achieves significantly higher AAR (+9.45\%) and MRR (+16.10\%) compared to the baseline DPLM while maintaining comparable designability. This highlights the importance of incorporating functional annotations to guide the design process.

\begin{figure} 
\includegraphics[width=0.99\linewidth]{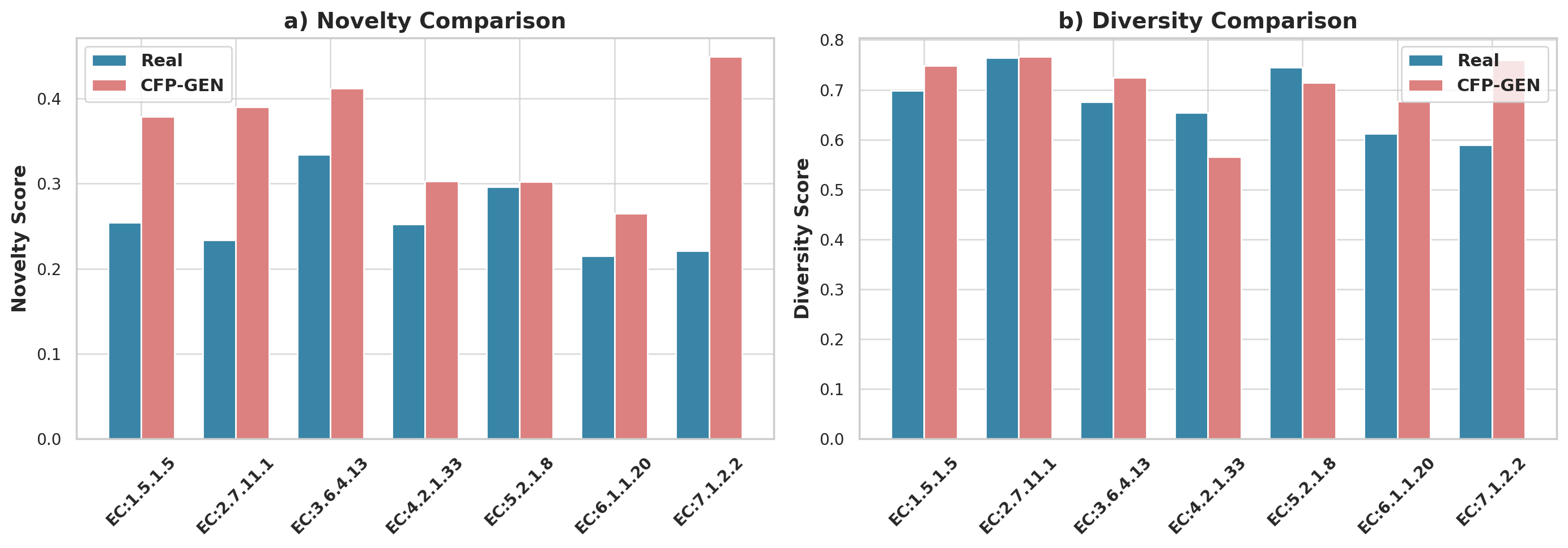}
\centering
 \caption{\textbf{Comparison of sequence novelty and diversity} between real and our designed proteins across 7 typical EC numbers from different enzyme families.}
\label{fig:novelty}
\end{figure}

\subsection{Multi-objective Protein Design}

We further demonstrate that using combinations of multimodal conditions enables the design of multi-functional proteins. Specifically, we consider six multi-catalytic enzymes as design targets, each annotated at least three EC numbers. For each enzyme, we construct three homologous sequences as validation set. Some generated proteins are visualized in Fig.~\ref{fig:proteins} to enhance interpretability. More introductions of these enzymes can be found in the Appendix \S\ref{subsec:enzymes}. 
We fine-tune \ourmodel~on the enzyme design dataset, where the homologous enzymes with same annotations are held out from the training set. We then compare our model against ZymCTRL by generating three sequences per enzyme, and the results are illustrated in Fig.~\ref{fig:multi-ec}. In particular, we compute the average success rate across all generated proteins, where a design is considered successful if all assigned EC numbers are present in the predicted labels, while also meeting the structural criteria of TM-score and pLDDT above 70. According to success rate and macro $\textbf{\textit{F}}_{\text{1}}$-score, \ourmodel~significantly outperforms ZymCTRL, highlighting the advantages of multimodal conditioning.

\subsection{Analysis of Novelty and Diversity}
A crucial aspect of \textit{de novo} protein design is the ability to generate novel and diverse proteins that do not exist in nature. To examine whether our model generates truly new proteins, we select 7 diverse EC numbers from different top-level categories: Oxidoreductases, Transferases, Hydrolases, Lyases, Isomerases, Ligases, and Translocases. For each class, \ourmodel~generates 30 sequences conditioned only on the EC number. We then compare with 30 real proteins from the enzyme validation set with the corresponding EC number. Novelty is computed by measuring how different (\ie, sequence identity after alignment) each generated sequence is from its most similar real protein in the training set, while diversity is computed by capturing how different the generated sequences are from the overall training set. To ensure both metrics are interpretable in the same direction, we subtract the scores from 1 (\ie, higher is better). The results are shown in Fig.~\ref{fig:novelty}. We observe that \ourmodel~consistently achieves higher sequence novelty across all 7 EC numbers, demonstrating its strong potential for \textit{de novo} protein design beyond simply replicating known sequences. Moreover, the generated sequences exhibit high intra-class diversity in 5 out of the 7 EC categories. These results suggest that the model has learned a more generalized representation, rather than overfitting to training examples. Additional analysis of \ourmodel~is presented in Appendix~\S\ref{sec:analysis}. 

\section{Conclusion and Future Work}

This work introduces a novel PLM, \ourmodel, for functional protein design with a multimodal conditioning mechanism. The core of \ourmodel~comprises three modules, each addressing functional annotations, sequence motifs, and backbone structures, respectively.  
Specifically, an Annotation-Guided Feature Modulation (AGFM) module is designed to flexibly and smoothly incorporate composable functional tags, such as GO terms, IPR domains, and EC numbers. A Residue-Controlled Functional Encoding (RCFE) module is introduced to encode crucial sequence motifs, facilitating scenarios where functional annotations are unavailable. Finally, a structure adapter is integrated into \ourmodel~without requiring additional training, ensuring coherence between function and structure.  
We evaluate \ourmodel~on three tasks: functional sequence generation, functional inverse folding, and multi-functional protein design, where it consistently outperforms existing approaches.

Although \ourmodel~represents an early attempt to jointly integrate multiple modalities for conditional protein generation, it has several limitations. First, it currently supports only a subset of GO, IPR, and EC labels, limiting its generalizability to broader functional categories. Second, beyond commonly used functional annotations, richer conditional inputs—such as physicochemical properties (\eg, hydrophobicity, charge, polarity)—are also essential to meet the requirements of bio-manufacturing applications. Moreover, sequence–structure co-design should be incorporated to enable end-to-end protein design. Future work will focus on scaling up the training datasets, enriching the scope of functional annotations, and further advancing unified models for more scalable and function-aware protein generation.

\section*{Impact Statement}

This work introduces a multimodal conditional protein design framework, enabling precise control over generated proteins by integrating diverse functional constraints. This approach has broad potential applications in enzyme engineering, drug discovery, and synthetic biology. We also provide a comprehensive benchmark of existing controllable PLMs, offering valuable insights for researchers utilizing these computational tools. 

From an ethical standpoint, our model is trained entirely on publicly available datasets (\eg, UniProtKB), ensuring reproducibility and transparency. However, potential biases may arise from unequal representation across protein families and functions, as public datasets often reflect human-curated biases. Addressing these imbalances requires continuous expansion of training data and careful validation on diverse proteins. Biosecurity risks must also be carefully considered. The ability of generative models to design novel proteins raises dual-use concerns, particularly in the context of biomedical applications. The safe use of biological generative models requires risk assessment, regulatory compliance and safeguards to prevent misuse while ensuring ethical and beneficial applications.

\section*{Acknowledgements}
This work was supported by the King Abdullah University of Science and Technology (KAUST) Office of Research Administration (ORA) under Award No REI/1/5234-01-01, REI/1/5414-01-01, REI/1/5289-01-01, REI/1/5404-01-01, REI/1/5992-01-01, URF/1/4663-01-01, Center of Excellence for Smart Health (KCSH), under award number 5932, and Center of Excellence for Generative AI, under award number 5940. For computer time, this research used Ibex managed by the Supercomputing Core Laboratory at KAUST in Thuwal, Saudi Arabia.

\nocite{langley00}

\bibliography{CFP-GEN.bib}
\bibliographystyle{icml2025}

\newpage
\appendix
\onecolumn
\section*{Supplementary Material of \ourmodel}

\section{Dataset Curation}
\label{subsec:dataset}

Existing PLMs for protein generation often focus on either unconditional sequence generation or single-condition controllable generation. For instance, ProteoGAN~\cite{kucera2022conditional} utilizes GO terms for functional control, ESM3~\cite{hayes2025simulating} leverages IPR domain information, and ZymCTRL~\cite{munsamy2022zymctrl} conditions on EC numbers for enzyme design. However, relying on single-condition constraints makes these datasets unsuitable for training a multimodal, multi-condition controllable PLM like \ourmodel, necessitating the curation of new datasets for functional protein design.

A protein is usually defined by three different modalities, \ie, function, sequence, and structure. These three aspects are generally considered to be highly interdependent and mutually constraining, as a protein’s sequence determines its structural fold, while both sequence and structure influence its functional properties. Given this intrinsic coupling, we annotate each protein from all three perspectives to ensure a comprehensive representation.
Specifically, we construct two specialized datasets to facilitate functional protein design: (1) A general protein design dataset with GO terms, IPR domains, functional sequence segments (motifs) and protein structures. (2) An enzyme design dataset specifically curated for generating catalytically active proteins by further adding EC number annotations.

\subsection{The General Protein Dataset.}

Gene Ontology (GO) terms provide a structured and comprehensive vocabulary for annotating proteins with respect to their molecular functions, biological processes, and cellular components, thereby offering a proper framework for understanding protein roles in various biological contexts. In this work, we focus primarily on molecular function annotations, which offer a more direct and interpretable characterization of each protein’s activity. In addition to GO annotations, we incorporate InterPro (IPR) numbers, which categorize proteins based on conserved domains and families, facilitating the identification of functional and structural relationships across diverse protein sequences. Combining GO and IPR labels allows for a deeper understanding of proteins from both global and local perspectives and thus facilitating the protein generation process.

To build the general protein dataset, we collect protein sequences from the manually curated Swiss-Prot subset of UniProtKB~\cite{uniprot2019uniprot}, ensuring high annotation reliability. To maintain balanced class distributions, we implement a filtering strategy that retains only GO and IPR terms with at least 100 annotated protein sequences, as suggested by ProteoGAN, to mitigate the impact of long-tail categories. Although this inevitably excludes some sequences from the training set, it has demonstrated the model’s capability to generalize across commonly used functional labels. Besides, 
our method primarily focuses on integrating different modalities as conditioning signals, enabling more comprehensive protein function modeling. We leave the addressing of long-tail data challenges in large-scale diffusion models to future iterations.

The final dataset contains 103,939 protein sequences covering 375 GO terms and 1,154 IPR domains. The distribution of GO and IPR annotations in our dataset is illustrated in Fig.~\ref{fig:dataset}. As shown in the GO annotation distribution, 30.6\% of GO categories contain 100–200 protein sequences, while only 6.4\% have more than 1,000 sequences. Similarly, in the IPR annotation distribution, 35.4\% of IPR categories consist of 100–200 sequences, whereas only 2.0\% contain more than 1,000 sequences.  Despite the remaining long-tail distribution in the dataset, we observe that the model generalizes well across different categories, as indicated by the macro $\textbf{$F_{\text{1}}$}$ score in Tab.~\ref{tab:sota_comparison}. 

Additionally, for each IPR number, we extract motif information from the annotated domain boundary positions. During both training and inference, we select the IPR domain most relevant to the given protein’s GO annotations (if multiple domains are present, we use their intersection) to define the functional sequence segments. These segments (motifs) are then used as sequence-level conditioning inputs to enhance the model’s ability to capture functionally relevant patterns. Furthermore, for each retained protein sequence, we retrieve its corresponding structural information from PDB and AlphaFold DB (AFDB) databases. Specifically, we extract the backbone atomic coordinates of 'N', 'CA', 'C', and 'O' and used them as structure condition. Our analysis reveals that the backbone structure plays a crucial role in determining protein function, proving the importance of incorporating structural features into PLMs.

To evaluate GO function, we construct a validation set by selecting 30 sequences per GO/IPR label, resulting in a subset of 8,309 sequences. The training dataset is then formed by holding out these sequences to ensure an unbiased evaluation.
For IPR function assignment, we further perform a 10-fold uniform downsampling of the GO validation set, yielding 831 sequences, to alleviate the computational overhead associated with InterProScan.

\subsection{The Enzyme Design Dataset.}

\begin{figure}[t]
\includegraphics[width=0.99\linewidth]{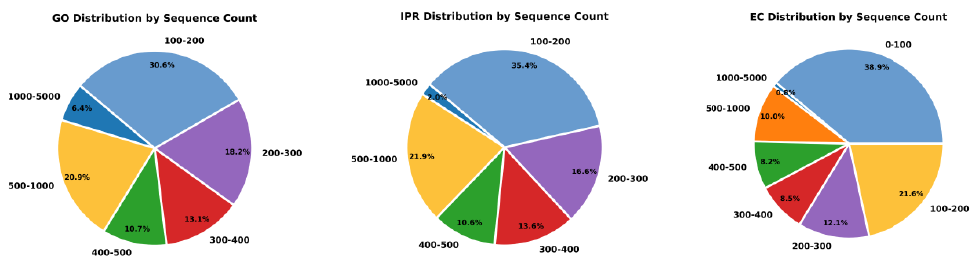}
\centering
\caption{ \textbf{Categorical distribution of functional annotations in the two datasets.} Most functional labels contain fewer than 500 sequences, highlighting the long-tail distribution issue. However, \ourmodel~exhibits strong generalization across these functions.}
\label{fig:dataset}
\end{figure}

Enzymes, as a critically important class of proteins with wide applications in biotechnology and industrial catalysis, requiring a specialized dataset for high-quality enzyme design. EC numbers (Enzyme Commission numbers) classify enzymes based on their catalytic activities. Linking protein sequences to specific biochemical reactions facilitates a better understanding of enzyme function. To construct a robust enzyme dataset, we leverage CARE~\cite{yang2024care}, a newly curated enzyme dataset comprising 185,995 sequences, each annotated with a level-4 EC number for precise functional clustering. To ensure compatibility with GO and IPR annotations, we extract the subset of CARE that overlaps with Swiss-Prot, yielding a dataset of 139,551 sequences spanning 661 EC labels. The distribution of EC categories is illustrated in Fig.~\ref{fig:dataset}.

In line with our approach for the General Protein Dataset, we also integrate sequence motifs from IPR annotations and structural information from PDB/AFDB databases. To enable rigorous evaluation, we construct a validation set by sampling 30 sequences per EC label, resulting in a high-quality evaluation set of 16,187 sequences, while the remaining data is allocated to the training set. Besides, multi-catalytic enzymes in Fig.~\ref{fig:multi-ec} are also held out from the training set to ensure a fair evaluation. As demonstrated in Tab.~\ref{tab:sota_comparison}, our model achieves catalytic function comparable to natural enzymes on this large-scale validation set, as assessed by the widely used CLEAN tool. 

\section{Evaluation Metrics}
\label{subsec:metrics}

We conduct a comprehensive evaluation of the generated proteins across multiple levels to assess their quality rigorously:

(1) Sequence-level evaluation: We measure the similarity between the generated sequences and natural protein sequences. Additionally, we analyze the inter-group distances by clustering sequences based on their functional labels, ensuring that functionally related sequences exhibit appropriate distribution.

(2) Function-level evaluation: We assess the functional consistency between the functions of generated sequences (\ie, estimated by SOTA function predictors) and their input prompts (\ie, derived from natural sequences). This is evaluated through multi-label classification metrics, ensuring that the predicted functions  align with their prompt functions.

(3) Structure-level evaluation: We evaluate the structural consistency of the generated sequences by predicting their 3D structures using ESMFold and comparing them with their corresponding PDB/AFDB structures. This ensures that the generated proteins not only maintain sequence-level fidelity but also exhibit high designability.

\subsection{Sequence Evaluation Metrics}

\begin{itemize}

\item \textbf{Maximum Mean Discrepancy (MMD)}: 
Maximum Mean Discrepancy (MMD) measures the distribution difference between generated protein sequences $P$ and their corresponding natural sequences $S$. Due to its efficiency and high expressive capacity, we adopt normalized Spectrum Mapping to obtain sequence distribution, as suggested by ProteoGAN. This method computes k-mer occurrences within a sequence, providing a sufficiently rich representation to effectively distinguish protein properties. Given a kernel function $k(\cdot, \cdot)$, MMD is computed as:
\begin{equation}
\text{MMD}^2 (S, P) = \mathbb{E}_{s,s' \sim S}[k(s, s')] + \mathbb{E}_{p,p' \sim P}[k(p, p')] - 2\mathbb{E}_{s\sim S, p\sim P}[k(s, p)]
\end{equation}
where $s, s' \in S$ are real sequences, and $p, p' \in P$ are generated sequences. A lower MMD value indicates better alignment between the generated and real distributions. For a linear kernel, the MMD computation simplifies to a direct Euclidean distance between the mean embeddings of the two distributions:
\begin{equation}
\text{MMD}_{\text{linear}}(S, P) = \|\mu_S - \mu_P\|, \quad\text{where,}\quad  \mu_S = \frac{1}{m} \sum_{i=1}^{m} s_i, \quad \mu_P = \frac{1}{n} \sum_{j=1}^{n} p_j
\end{equation}
represent the mean embeddings of the real and generated sequences, respectively, and $\|\cdot\|$ denotes the Euclidean norm.

\item \textbf{Gaussian Kernel MMD (MMD-G)}: 
For an RBF (Gaussian) kernel, the MMD computation incorporates the radial basis function (RBF) to capture higher-order distributional differences:
\begin{equation}
\text{MMD}_{\text{Gaussian}}^2 (S, P) = \frac{1}{m^2} \sum_{i,j} k(s_i, s_j) + \frac{1}{n^2} \sum_{i,j} k(p_i, p_j) - \frac{2}{mn} \sum_{i,j} k(s_i, p_j)
\end{equation}
where the Gaussian kernel function is defined as:
\begin{equation}
k(x, y) = \exp\left(-\gamma \|x - y\|^2\right)
\end{equation}
with $\gamma$ typically determined via a median heuristic approach to adaptively select an appropriate kernel bandwidth, \ie,
\begin{equation} 
\sigma = \text{median}({ | x_i - x_j | }_{i,j}), \quad \gamma = \frac{1}{2\sigma^2} 
\end{equation}
where $\sigma$ is the median of all pairwise Euclidean distances between embeddings from both groups.

By considering both the linear kernel MMD, which captures global differences, and the Gaussian kernel MMD, which captures higher-order distributional variations, we ensure a robust evaluation of the similarity between real and generated protein distributions.

\item \textbf{Mean Reciprocal Rank (MRR)}:
Mean Reciprocal Rank (MRR) evaluates how well the generated sequences $P$ match the real sequences $S$ in a function-aware manner based on linear MMD distance.

We define $S = \{S_c\}_{c=1}^{C}$ as a set of real protein sequences grouped by their function label $c$, where $C$ is the total number of function labels (\ie, GO/IPR/EC annotations). Similarly, let $P = \{P_c\}_{c=1}^{C}$ be the corresponding generated protein sequences for each functional group. The goal is to assess whether the generated sequences for each function closely match the real ones. The MRR metric is defined as:
\begin{equation}
\text{MRR}(S, P) = \frac{1}{C} \sum_{c=1}^{C} \frac{1}{\text{rank}_S(\text{MMD}(S_c, P_c))}
\end{equation}
where $\text{rank}_S(\text{MMD}(S_c, P_c))$ represents the ranking position of the MMD score between the real sequences $S_c$ and generated sequences $P_c$. 
Specifically, for each function label pair $(c, c')$, we compute $\text{MMD}(S_c, P_{c'})$, measuring the distributional difference between the real sequences of function $c$ and the generated sequences of function $c'$. The ranking of $\text{MMD}(S_c, P_c)$ is then determined by sorting these MMD values across all function pairs.

MRR ensures that the evaluation considers not only how well each function’s generated sequences match their corresponding real sequences but also their relative distinction from other functional groups. A higher MRR indicates better functional alignment between the real and generated distributions. When $\text{MRR}(S, P) = 1$, it signifies that for each functional class, the generated sequences are the closest in distribution to their corresponding real sequences, demonstrating high-quality protein generation.
\end{itemize}

\subsection{Function Evaluation Metrics}
\begin{itemize}

    \item \textbf{Macro $\textbf{\textit{F}}_{\text{1}}$-score}: 
    Macro $\textbf{\textit{F}}_{\text{1}}$-score computes the per-class $\textbf{\textit{F}}_{\text{1}}$-scores, treating all classes equally, which is denoted as:
    \begin{equation}
    \text{macro } \textbf{\textit{F}}_{\text{1}} = \frac{1}{C} \sum_{i=1}^{C} \frac{2 \cdot \text{Precision}_i \cdot \text{Recall}_i}{\text{Precision}_i + \text{Recall}_i}
    \end{equation}
    where $C$ is the number of classes, and $\text{Precision}_i$, $\text{Recall}_i$ are the precision and recall for class $i$. This metric is particularly useful for evaluating performance across imbalanced datasets, as it ensures that smaller classes contribute equally to the final score.

    \item \textbf{Micro $\textbf{\textit{F}}_{\text{1}}$-score}: 
    In contrast, micro $\textbf{\textit{F}}_{\text{1}}$-score aggregates the counts of true positives, false positives, and false negatives across all classes before computing a single $\textbf{\textit{F}}_{\text{1}}$-score:
    \begin{equation}
    \text{micro } \textbf{\textit{F}}_{\text{1}} = \frac{2 \cdot \text{micro Precision} \cdot \text{micro Recall}}{\text{micro Precision} + \text{micro Recall}}
    \end{equation}
    where micro precision and micro recall are computed as:
    \begin{equation}
    \text{micro Precision} = \frac{\sum_{i=1}^{C} TP_i}{\sum_{i=1}^{C} (TP_i + FP_i)}, \quad     \text{micro Recall} = \frac{\sum_{i=1}^{C} TP_i}{\sum_{i=1}^{C} (TP_i + FN_i)}
    \end{equation}
    where $TP_i$, $FP_i$, and $FN_i$ represent true positives, false positives, and false negatives for class $i$, respectively. Unlike macro $\textbf{\textit{F}}_{\text{1}}$-score, micro $\textbf{\textit{F}}_{\text{1}}$-score gives more weight to high-frequency classes.

    \item \textbf{Macro AUPR (Area Under the Precision-Recall Curve)}:
    Macro AUPR computes the area under the precision-recall curve for each class and averages them:
    \begin{equation}
    \text{macro AUPR} = \frac{1}{C} \sum_{i=1}^{C} \text{AUPR}_i
    \end{equation}
    where $\text{AUPR}_i$ is the precision-recall area for class $i$. This metric is particularly valuable in scenarios with highly imbalanced datasets, where precision-recall trade-offs are more informative than ROC-based metrics.

    \item \textbf{Macro AUC (Area Under the Receiver Operating Characteristic Curve)}:
    Macro AUC measures classification performance by averaging the AUC values across all classes:
    \begin{equation}
    \text{macro AUC} = \frac{1}{C} \sum_{i=1}^{C} \text{AUC}_i
    \end{equation}
    where $\text{AUC}_i$ is the ROC-AUC for class $i$. Higher values indicate better model discrimination between positive and negative instances.

   \item \textbf{$\textbf{\textit{F}}_{\text{max}}$-score}: 
In Tab.~\ref{tab:sota_comparison}, we evaluate function scores using standard multi-label metrics, with confidence thresholds of 0.1 for GO and 0.01 for EC. To account for the hierarchical structure of GO functions, ancestor GO terms are also incorporated in evaluation. Additionally, in Tab.~\ref{tab:inverse}, to align with DeepGO-SE, we also consider a $\textbf{\textit{F}}_{\text{max}}$-score computed at the sequence level. Given a confidence threshold $\tau$ (ranging from 0.1 to 1.0 in increments of 0.01), we define:
- $TP_{\tau}$: The number of correctly predicted GO terms present in the ground truth set of each sequence.
- $FP_{\tau}$: The number of predicted GO terms that do not exist in the ground truth of each sequence.
- $FN_{\tau}$: The number of ground truth GO terms that were not predicted. Using these, $\text{Precision}_{\tau}$ and $\text{Recall}_{\tau}$ at threshold $\tau$ are calculated as:
\begin{equation}
\text{Precision}_{\tau} = \frac{TP_{\tau}}{TP_{\tau} + FP_{\tau}}, \quad
\text{Recall}_{\tau} = \frac{TP_{\tau}}{TP_{\tau} + FN_{\tau}}
\end{equation}

The $\textbf{\textit{F}}_{\text{max}}$-score is then obtained by maximizing the $\textbf{\textit{F}}_{\text{1}}$-score over a range of confidence thresholds:
\begin{equation}
\textbf{\textit{F}}_{\text{max}} = \max_{\tau} \left\{ \frac{2 \cdot \text{Precision}_{\tau} \cdot \text{Recall}_{\tau}}{\text{Precision}_{\tau} + \text{Recall}_{\tau}} \right\}
\end{equation}

Unlike macro $\textbf{\textit{F}}_{\text{1}}$-score, which averages per-class performance, $\textbf{\textit{F}}_{\text{max}}$ focuses on per-sequence predictions.

\end{itemize}

\subsection{Structure Evaluation Metrics}

\begin{itemize}

    \item \textbf{Amino Acid Recovery (AAR)}:  
    AAR measures how well a generated protein sequence $\mathbf{x'}$ matches a reference sequence $\mathbf{x}$, given the protein structures as model inputs. Let $\mathbf{x'} = (x'_1, x'_2, \dots, x'_L)$ and  $\mathbf{x} = (x_1, x_2, \dots, x_L)$ denote the generated sequence and the reference sequence of length $L$. AAR is then defined as:  
\begin{equation}
\text{AAR} = \frac{1}{L} \sum_{i=1}^{L} \delta(x'_i, x_i), \quad
\delta(x'_i, x_i) =
\begin{cases} 
1, & \text{if } x'_i = x_i \\
0, & \text{otherwise}
\end{cases}
\end{equation}
The function $\delta(x'_i, x_i)$ is the indicator function. A higher $\text{AAR}$ value indicates greater sequence fidelity to the reference, meaning the generated sequence closely resembles the original target sequence at the residue level.

    \item \textbf{TM-score (scTM)}:  
    scTM globally evaluates the structural similarity between a generated protein sequence $\mathbf{x'}$ and a reference protein sequence $\mathbf{x}$, considering the backbone structure $c_\text{str}$ as a conditioning factor. Let $\mathbf{F}(x)$ and $\mathbf{F}(x')$ be the 3D structures predicted for sequences $\mathbf{x}$ and $\mathbf{x'}$ respectively (using ESMFold in our case), then scTM is defined as:
    \begin{equation}
    \text{scTM} = \text{TM-score}(\mathbf{F}(x'), \mathbf{F}(x) \mid c_\text{str})
    \end{equation}
 Higher scTM scores indicate better structural consistency under backbone structure guidance.

    \item \textbf{Predicted Local Distance Difference Test (pLDDT)}:  
    pLDDT quantifies the confidence of a predicted protein structure by assessing the local agreement of residue positions. Given a predicted structure $\mathbf{F}(x')$ with $L$ residues, the pLDDT score is computed as:
    \begin{equation}
    \text{pLDDT} = \frac{1}{L} \sum_{i=1}^{L} \text{pLDDT}_i
    \end{equation}
    where $\text{pLDDT}_i$ is the per-residue confidence score of $x_i \in \{x'_1, x'_2, \dots, x'_L\}$ derived from the folding model. A higher pLDDT score indicates greater reliability in the predicted structure.

\end{itemize}

\begin{figure}[t]
\includegraphics[width=0.99\linewidth]{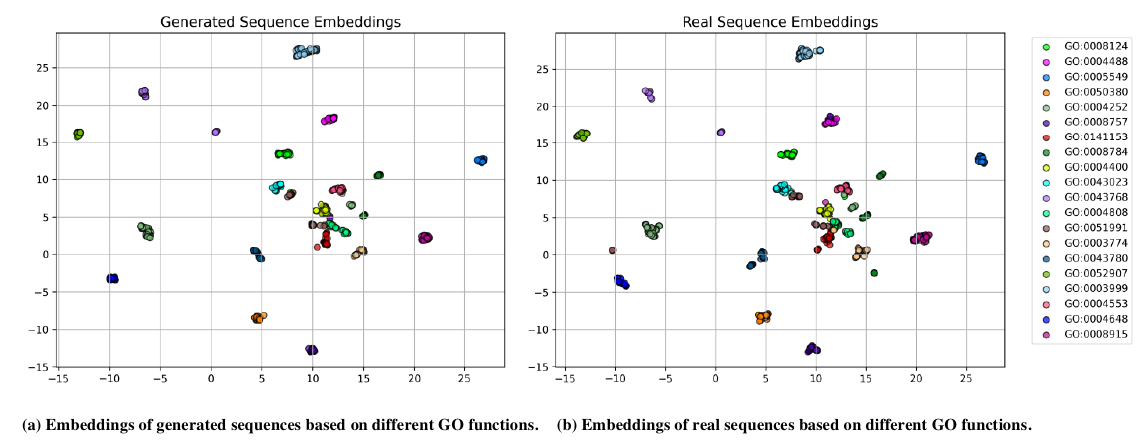}
\centering
\caption{ \textbf{UMAP visualization of sequence embeddings conditioned on GO functions.} The generated sequences from \ourmodel~exhibit a highly similar distribution to natural sequences, demonstrating the strong expressiveness and functional consistency of our approach.}
\label{fig:umap}
\end{figure}

\section{Extensive Analysis of \ourmodel}
\label{sec:analysis}
\subsection{Analysis of Generated Protein Embeddings}
To evaluate the alignment between generated and real protein sequences, we compare their embedding distributions across 20 randomly selected GO terms, with 30 sequences per GO term for both real and generated data. The embeddings are obtained using Spectrum Mapping, a highly expressive sequence representation method, and visualized via UMAP for dimensionality reduction. As shown in Fig.~\ref{fig:umap}, the distributions of generated sequences closely align with those of real sequences, with each functional category maintaining a consistent spatial position across both plots. The inter-cluster distances remain well-preserved, indicating that proteins with different functions occupy distinct embedding spaces, while the intra-class compactness suggests that functional specificity is effectively captured. Additionally, the model generalizes well across diverse functional spaces, preserving structured and biologically meaningful separations. Minor overlaps in some central regions suggest that certain sequences may exhibit multiple GO functions, reflecting the inherent complexity of protein function annotation. The overall results demonstrate that \ourmodel~leverages Spectrum Mapping embeddings to effectively preserve the functional landscape of protein sequences, ensuring both functional coherence and class separability in the learned representation space.

\subsection{Mitigation of Mode Collapse.}
One common issue with PLMs is mode collapse in the generated sequences. To evaluate whether \ourmodel~alleviates this problem, we analyzed the frequency of repeated $n$-gram patterns ($n = 2, 3, 4, 5, 6$) in sequences generated under GO-conditioned prompts, as reported in Tab.\ref{tab:sota_comparison}. Sequences generated by DPLM are used as a reference, and real protein sequences from the validation set serve as a positive control.

As shown in Tab.~\ref{tab:ngram_overlap}, \ourmodel~produces a similar number of 2-grams to real proteins, while significantly reducing the number of longer repetitive n-grams, especially 4-gram to 6-gram patterns. Notably, the more functional conditions (\eg, GO, IPR, Motif) are provided, the fewer repetitive patterns appear in the output, indicating better sequence quality and reduced mode collapse. These results provide strong evidence that \ourmodel~effectively alleviates the mode collapse issue observed in mainstream PLMs.

\begin{table}[h]
\centering
\vspace{-4mm}
\caption{Comparison of $n$-gram counts across different methods.}
\label{tab:ngram_overlap}
\begin{tabular}{lccccc}
\toprule
\textbf{Method} & \textbf{2-gram} & \textbf{3-gram} & \textbf{4-gram} & \textbf{5-gram} & \textbf{6-gram} \\
\midrule
Positive Control                  & 404 & 164 & 0   & 0  & 0  \\
DPLM                              & 315 & 462 & 104 & 46 & 26 \\
\ourmodel~(\textit{w/ GO})                   & 363 & 351 & 15  & 9  & 8  \\
\ourmodel~(\textit{w/ GO and IPR})           & 365 & 332 & 9   & 5  & 4  \\
\ourmodel~(\textit{w/ GO, IPR and Motif})    & 377 & 336 & 4   & 1  & 1  \\
\bottomrule
\end{tabular}
\end{table}

\section{Hyperparameter Details}
\label{subsec:details}
As introduced in the main paper, unless otherwise specified, most of the learning strategies and hyperparameters of the diffusion model remain consistent with those of DPLM. The batch size is set to 1 million tokens, and training is conducted on 8 NVIDIA A100 GPUs for around 72 hours of each stage. The AdamW optimizer is employed with a maximum learning rate of 0.00004. During inference, we allow the model to perform 100 sampling steps, following the DPLM conditional generation, with sequence length varying from 200 to 400. The total model size of \ourmodel~is 1.48B parameters, excluding the GVP-transformer structure encoder.

\begin{figure}[t]
\includegraphics[width=0.99\linewidth]{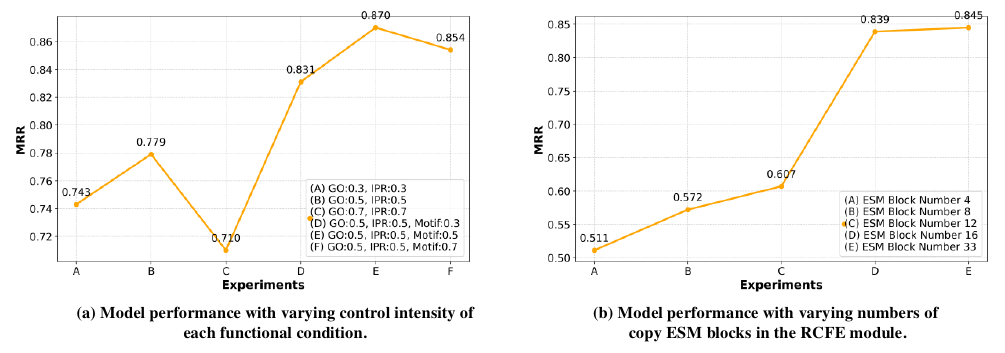}
\centering
\caption{ \textbf{Hyperparameter Analysis of \ourmodel.} We examine the the impact of control intensity and the number of copy ESM blocks.}
\label{fig:hyper}
\end{figure}

During training, we adjust $\gamma(c)$ in Eq.~\ref{eq:gamma} by applying dropout to each conditioning input. The probability of randomly dropping out each condition and its corresponding impact on model performance is shown in Fig.~\ref{fig:hyper} (a). MRR is used to present the performance as we found this metric is more robust and expressive. The results indicate that setting the dropout probability to 0.5 for each condition achieves the best performance. Additionally, we investigate the optimal number of ESM block copies in the RCFE module. We compare the model performance conditioning on only the sequence motifs. As shown in Fig.~\ref{fig:hyper} (b), setting the number of blocks to 16 achieves the best trade-off between performance and model complexity.

\section{Implementation of Existing PLMs}
\label{subsec:plms}

For ProGen2, we provide the model with GO terms along with the first 30 residues of the real sequence as prompts. When evaluating IPR/EC functions, only the first residues are used, as ProGen2 does not support EC/IPR annotations.

For ProteoGAN, we directly input the GO terms of the real sequence. However, since ProteoGAN only supports 50 predefined GO terms, for GO terms not included in ProteoGAN’s vocabulary, we attempt to map them to their closest ancestor terms that are supported. If no suitable ancestor is found, the sequence is ignored.

For discrete diffusion models such as DPLM, we provide the model with functional motifs (30 residues) and task it with performing sequence inpainting to reconstruct the missing residues.

For ESM3, we input both the IPR domain descriptions and their start-end positions, along with 30 residues to initialize the sequence generation process.

For ZymCTRL, only a single EC number per sequence is provided, as the model does not support multi-label inputs. If an EC number is not supported by ZymCTRL, we exclude the sequence from evaluation to ensure a fair comparison.

\section{Introduction of Multi-catalytic Enzymes}
\label{subsec:enzymes}

\begin{itemize}
\item 5$^{\prime}$/3$^{\prime}$-Nucleotidase SurE (\textbf{EC Numbers: 3.1.3.5, 3.1.3.6, 3.6.1.11}): This nucleotidase has {broad substrate specificity}, dephosphorylating various {ribo- and deoxyribonucleoside 5$^{\prime}$-monophosphates} and {ribonucleoside 3$^{\prime}$-monophosphates}, with the highest affinity for {3$^{\prime}$-AMP}. It also hydrolyzes {polyphosphate} ({exopolyphosphatase activity}), preferentially targeting short-chain-length substrates ({P20-25}). This enzyme is potentially involved in the {regulation of dNTP and NTP pools}, as well as the {turnover of 3$^{\prime}$-mononucleotides}, which are produced by various intracellular RNases ({T1, T2, and F}) during RNA degradation.
	\item Fatty Acid Oxidation Complex Subunit Alpha (\textbf{EC Numbers: 1.1.1.35, 4.2.1.17, 5.1.2.3, 5.3.3.8}): This enzyme plays a crucial role in the {fatty acid $\beta$-oxidation cycle}, enabling both {aerobic and anaerobic degradation} of long-chain fatty acids. It catalyzes the conversion of {enoyl-CoA to 3-oxoacyl-CoA} via {L-3-hydroxyacyl-CoA} and can also process {D-3-hydroxyacyl-CoA} and {cis-3-enoyl-CoA} as substrates. This pathway is fundamental for {cellular energy production} and lipid metabolism.  
	\item Geranylgeranyl Pyrophosphate Synthase (\textbf{EC Numbers: 2.5.1.1, 2.5.1.10, 2.5.1.29}): This enzyme catalyzes the synthesis of {geranylgeranyl pyrophosphate (GGPP)} from {farnesyl pyrophosphate and isopentenyl pyrophosphate}, initiating the {janthitremane biosynthesis pathway}. It participates in multiple steps involving {prenylation, oxidation, and cyclization}, ultimately leading to the formation of complex {indole diterpenes} such as {paspaline and shearinine A}. These compounds exhibit diverse bioactivities.  
	\item Siroheme Synthase (\textbf{EC Numbers: 1.3.1.76, 2.1.1.107, 4.99.1.4}): This multifunctional enzyme catalyzes a series of key reactions in the {siroheme biosynthesis pathway}. It first facilitates the {S-adenosylmethionine (SAM)-dependent methylation} of {uroporphyrinogen III} at {C-2 and C-7}, forming {precorrin-2} via {precorrin-1}. It then catalyzes the {NAD-dependent ring dehydrogenation} of {precorrin-2} to produce {sirohydrochlorin}. Finally, it promotes the {ferrochelation of sirohydrochlorin}, leading to the formation of {siroheme}, an essential cofactor in {sulfite and nitrite reductases}.
\item Putative Fatty Acid Oxidation Complex Trifunctional Enzyme (\textbf{EC Numbers: 1.1.1.35, 4.2.1.17, 5.3.3.8}): This enzyme exhibits multiple catalytic activities involved in {fatty acid metabolism}, including {3-hydroxyacyl-CoA dehydrogenase activity}, {delta(3)-delta(2)-enoyl-CoA isomerase activity}, and {enoyl-CoA hydratase activity}. It also possesses {NAD$^+$-binding} capacity and plays a role in {fatty acid catabolic processes}. This enzyme is essential in the {$\beta$-oxidation pathway}, facilitating the conversion of fatty acids into energy.
 \item Purine Nucleoside Phosphorylase RC0672 (\textbf{EC Numbers: 2.4.2.1, 2.4.2.28, 3.5.4.4}): This purine nucleoside enzyme catalyzes the {phosphorolysis of adenosine and inosine nucleosides}, producing {D-ribose 1-phosphate} and the respective free bases, {adenine and hypoxanthine}. It also catalyzes the {phosphorolysis of S-methyl-5'-thioadenosine} into {adenine and S-methyl-5-thio-$\alpha$-D-ribose 1-phosphate}. Additionally, it exhibits {adenosine deaminase activity}, contributing to purine metabolism.

\end{itemize}

\section{Visualization of Generated Proteins}
\begin{figure}[t]
\includegraphics[width=0.95\linewidth]{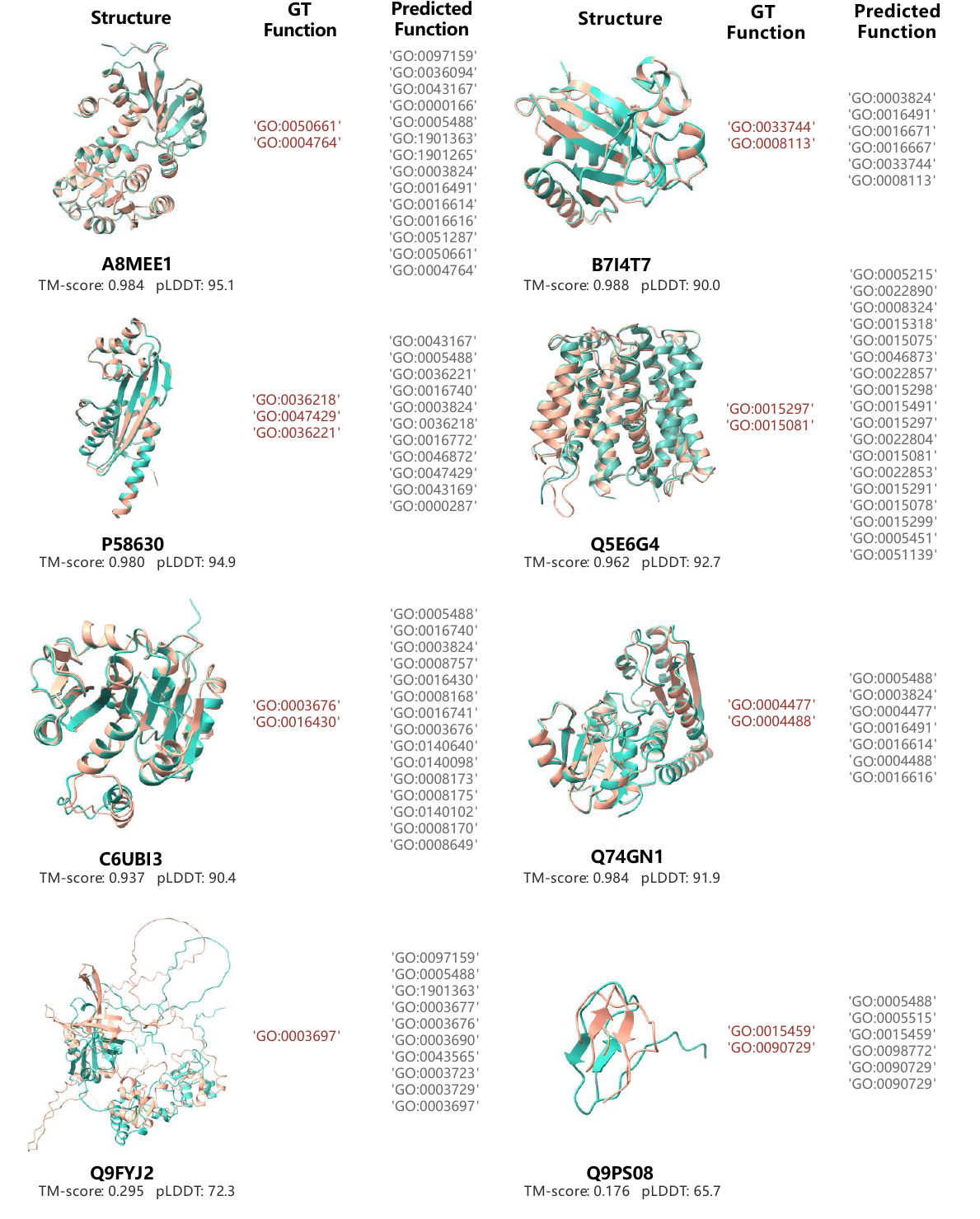}
\centering
 \vspace{-2mm}
\caption{ \textbf{Examples of generated functional proteins.} We compare the generated proteins and their predicted functions with corresponding PDB/AFDB structures and functions. These results demonstrate the effectiveness and functional consistency of \ourmodel.}
\label{fig:visual_go}
\end{figure}

In Fig.~\ref{fig:visual_go}, we present additional visualizations of generated proteins produced by \ourmodel~under multimodal conditions, specifically GO terms and backbone atomic coordinates. The structures are predicted using ESMFold, applied to the generated sequences. The reference PDB/AFDB structures, from which the backbone constraints are derived, are shown in green, while the structures of the designed proteins are shown in red. The real functions of the reference proteins are labeled in red, whereas the predicted functions of the generated proteins, inferred using DeepGO-SE, are labeled in gray. 

Several key observations can be drawn from these results. First, most generated proteins exhibit high structural fidelity, as indicated by TM-scores and pLDDT values exceeding 90, demonstrating that \ourmodel~can effectively design structurally stable proteins. Second, the majority of prompted functions are successfully recovered. Since DeepGO-SE also predicts ancestor GO terms, the predicted functions often include more terms than the input prompt. Here, we display only those with a confidence score greater than 0.3, further confirming \ourmodel’s ability to design functionally relevant proteins. Third, some failure cases were also identified. For instance, UniProt ID Q9FYJ2 contains long, flexible loop regions extending far from the core fold. These regions are typically associated with intrinsically disordered regions (IDRs), which lack a single well-defined conformation. As a result, the predicted structure exhibits high deviations in loop regions compared to the ground truth structure. Similarly, UniProt ID Q9PS08 corresponds to an extremely short protein (around 30 residues), which poses significant challenges for both generative and folding models, as such short sequences are underrepresented in large protein datasets. Consequently, these proteins exhibit low TM-scores, highlighting the inherent difficulties in designing ultra-short protein sequences. These findings demonstrate the strengths and limitations of \ourmodel~in functional protein generation, emphasizing its effectiveness in generating proteins with both high-fidelity structures and functional relevance while also highlighting areas for future improvements to enhance model generalization.

\end{document}